\documentclass[sigconf]{acmart}

\usepackage{amsmath}
\usepackage{algorithmic}
\usepackage{graphicx}
\usepackage{adjustbox}
\usepackage{textcomp}
\usepackage{xcolor}
\usepackage{tikz}
\usepackage{tikz-qtree}
\usetikzlibrary{mindmap,trees}
\usetikzlibrary{arrows}
\usetikzlibrary{calc}
\usepackage{etoolbox}

\usepackage{graphicx}
\usepackage[font={bf,it,small},labelsep=colon]{caption}
\usepackage{subcaption}
\usepackage{lipsum} 
\usepackage{caption}
\usepackage{setspace}
\usepackage{makecell}
\AtBeginEnvironment{quote}{\singlespace\vspace{-\topsep}\small}
\AtEndEnvironment{quote}{\vspace{-\topsep}\endsinglespace}
\usepackage{todonotes}
\usepackage{xspace}
\usepackage{color}
\usepackage{enumitem}
\usepackage{flushend}
\usepackage{comment}
\usetikzlibrary{tikzmark}
\usepackage{multicol}
\setcopyright{none}
\renewcommand\footnotetextcopyrightpermission[1]{} 

\newcommand{\ignore}[1]{}

\def\BibTeX{{\rm B\kern-.05em{\sc i\kern-.025em b}\kern-.08em
    T\kern-.1667em\lower.7ex\hbox{E}\kern-.125emX}}
\AtBeginDocument{%
  \providecommand\BibTeX{{%
    \normalfont B\kern-0.5em{\scshape i\kern-0.25em b}\kern-0.8em\TeX}}}

\begin{document}
\sloppy
\title{Anomaly Detection through Transfer Learning in Agriculture and Manufacturing IoT Systems\\}


\author{Mustafa Abdallah}
\affiliation{%
  \institution{Purdue University}
  \streetaddress{Electrical and Computer Engineering}
  \city{West Lafayette, IN}
  \country{USA}}
\email{abdalla0@purdue.edu}

\author{Wo Jae Lee}
\affiliation{%
  \institution{Purdue University}
  \streetaddress{Environmental and Ecological Engineering}
  \city{West Lafayette, IN}
  \country{USA}}
  
\author{Nithin Raghunathan}
\affiliation{%
  \institution{Purdue University}
  \streetaddress{Electrical and Computer Engineering}
  \city{West Lafayette, IN}
  \country{USA}}
  
\author{Charilaos Mousoulis}
\affiliation{%
  \institution{Purdue University}
  \streetaddress{Electrical and Computer Engineering}
  \city{West Lafayette, IN}
  \country{USA}}

\author{John W. Sutherland}
\affiliation{%
  \institution{Purdue University}
  \streetaddress{Environmental and Ecological Engineering}
  \city{West Lafayette, IN}
  \country{USA}}
  
\author{Saurabh Bagchi}
\affiliation{%
  \institution{Purdue University}
  \streetaddress{Electrical and Computer Engineering}
  \city{West Lafayette, IN}
  \country{USA}}  
\email{sbagchi@purdue.edu}  
\begin{abstract}

IoT systems have been facing increasingly sophisticated technical problems
due to the growing complexity of these systems and their fast deployment
practices. Consequently, IoT managers have to judiciously detect failures (anomalies)
in order to reduce their cyber risk and operational cost. While there is a rich literature on anomaly detection in many
IoT-based systems, there is no existing work that documents the use
of ML models for anomaly detection in digital agriculture and in
smart manufacturing systems. These two application domains pose
certain salient technical challenges. In agriculture the data is often sparse, due
to the vast areas of farms and the requirement to keep the cost
of monitoring low. Second, in both domains, there are multiple
types of sensors with varying capabilities and
costs. The sensor data characteristics change with
the operating point of the environment or machines, such as, the RPM of the motor.
The inferencing and the anomaly detection processes therefore
have to be calibrated for the operating point.

In this paper, we analyze data from sensors deployed in an agricultural farm with data from seven different kinds of sensors, and from an advanced manufacturing testbed with vibration sensors. We evaluate the performance of ARIMA and LSTM models for predicting the time series of sensor data. Then, considering the sparse data from one kind of sensor, we perform transfer learning from a high data rate sensor. We then perform anomaly detection using the predicted sensor data. Taken together, we show how in these two application domains, predictive failure classification can be achieved, thus paving the way for predictive maintenance. 
\end{abstract}



\maketitle

\section{Introduction}
The Internet of Things (IoT) \cite{atzori2010internet,gubbi2013internet} is a network of internet-connected objects able to collect and exchange data using embedded sensors. Any stand-alone internet-connected device that can be monitored and/or controlled from a remote location is considered an IoT device. Facing increasingly sophisticated technical problems due to the growing complexity of IoT systems and fast deployment practices, IoT managers have to judiciously detect failures (anomalies) in order to reduce their cyber risk and operational cost. In this context, anomaly detection is an important problem that has been researched within diverse research areas and application domains.

Anomaly detection is the identification of rare items, events or observations which raise suspicions by differing significantly from the majority of the data. The research on anomaly detection begins from the hypothesis that exploitative behavior is quantitatively distinct from normal system behavior (see the seminal paper \cite{denning1987intrusion}) which formed the basis of most modern anomaly detectors (e.g., \cite{ertoz2004minds,lee2001real}). Also, many researchers have used the presence of anomalous activity as a trigger to detect a security attack is in progress or evidence of malicious behaviour (see the survey \cite{jyothsna2011review}). 

In the IoT space, several anomaly detection techniques have been developed that leverage characteristics of specific application domains \cite{chandola2009anomaly,sabahi2008intrusion}. For example, Density-based techniques (e.g., (k-nearest neighbor \cite{knorr2000distance}, and local outlier factor \cite{breunig2000lof})) and one-class support vector machines \cite{platt1999estimating} detection techniques have been used in the literature. Most of the existing work has relied on these models for anomaly detection and failure detection in IoT-based systems \cite{7474197,shahzad2016energy,mitchell2014survey,chatterjee2020context}. 

While there is a rich literature on anomaly detection in many IoT-based systems, there is no existing work that documents the use of ML models for anomaly detection in digital agriculture\footnote{The term "digital agriculture" loosely is used to denote the use of modern technology, including prominently, IoT and wireless communication in agriculture practices.} and in smart manufacturing systems. These two application domains pose certain salient technical challenges for the use of ML-based models for anomaly detection. In agriculture the data is often sparse, due to the vast areas of farms and the requirement to keep the cost of monitoring low. Second, in both domains, there are multiple types of sensors concurrently generating data about the same (or overlapping) events. These sensors are of varying capabilities and costs. In manufacturing, the sensor data characteristics change with the operating point of the machines, such as, the RPM of the motor. The inferencing and the anomaly detection processes therefore have to be calibrated for the operating point. Thus, we need case studies of anomaly detection deployments on these two IoT-based systems --- the need for such deployments and resultant analyses have been made separately for digital agriculture \cite{8710531} and for smart manufacturing systems~\cite{thomas2018minerva,scime2018anomaly} (also see the survey paper \cite{wang2018deep} on the usage and challenges of deep learning in smart manufacturing systems).

There is also important economic impetus for this kind of deployment and analysis. In a smart manufacturing system, various sensors (e.g., vibration, ultrasonic, pressure sensors) are applied for process control, automation, production plans, and equipment maintenance. For example, in equipment maintenance, the condition of operating equipment is continuously monitored using proxy measures (e.g., vibration and sound) to prevent unplanned downtime and to save maintenance costs \cite{lee2019mfg1}. Thus, the data from these sensors can be analyzed in a streaming, real-time manner to fill a critical role in predictive maintenance tasks, through the anomaly detection process~\cite{garcia2006simap,kroll2014system,de2018anomaly}. In the area of digital agriculture, for large farms, it is costly for humans to manually intervene in data collection (say, to change a failing sensor) or mitigation actions (say, to change the position of an irrigation machinery) and thus highly accurate anomaly detection can cut down on these labor costs.  
Thus, we propose our anomaly detection technique for smart agriculture (i.e., agriculture systems which is based on IoT and cloud computing~\cite{tongke2013smart}) and smart manufacturing systems~\cite{kusiak2018smart}. 

Two notable exceptions to the lack of prior work in these domains are the recent works \cite{christiansen2016deepanomaly,lee2019mfg2}. 
In~\cite{christiansen2016deepanomaly}, the authors used deep learning and anomaly detection to exploit the homogeneous characteristics of a field to perform anomaly detection on obstacles. However, it does not address anomaly detection of agriculture sensor readings due to weather conditions or sensor malfunction, that we consider here. In \cite{lee2019mfg2}, the authors proposed a kernel principal component analysis (KPCA)-based anomaly detection system to detect a cutting tool failure in a machining process. However, they did not address the domain-specific challenges introduced above.

\noindent {\bf Our Contribution}: \\
In this paper, we perform anomaly detection by considering machine learning (ML) models. We specialize the model to the constraints of the sensors respectively in agricultural and manufacturing settings.
We design a temporal anomaly detection technique and 
efficient defect-type classificaton technique.

In such large-scale IoT systems, the data are collected from different sensors via intermediate data collection points and finally aggregated to a server to further store, process, and perform useful data-analytics on the sensor readings \cite{marjani2017big,he2017multitier}. We propose a \emph{temporal anomaly detection} model, in which the temporal relationships between the readings of the sensors are captured via a time-series prediction model. Specifically, we consider two classes of time-series prediction models which are \textit{\bf Autoregressive Integrated Moving Average model (ARIMA)} and \textit{\bf Long Short-Term memory (LSTM)}. These models are used to predict the expected future samples in certain time-frame given the near history of the readings. We first test our models on real data collected from both deployed agricultural sensors and deployed manufacturing sensors to detect anomalous data readings. We then analyze the performance of our models, and compare the algorithms of ARIMA and LSTM time-series predictors.

 We observe that LSTM models gave better performance, in terms of Root Mean Squared Error (RMSE) on the test samples. In this context, LSTM performance is better (i.e., its RMSE is less) in six out of seven agricultural sensors compared to ARIMA model.

 Another problem in this domain is the prediction from models using sparse data, which is often the case because of limitations of the sensors or the cost of collecting data (for instance, in our manufacturing application domain, the MEMS sensors has low sampling frequency). One mitigating factor is that plentiful data may exist in a slightly different context, such as, from a different kind of sensor on the same equipment or the equipment being operated under a somewhat different operating condition in a different facility (such as, a different RPM). Thus, the interesting research question in this context is: can we use a model trained on data from a different kind of sensor (such as, a piezoelectric sensor, which has a high sampling frequency) to perform anomaly detection on data from a different kind of sensor (such as, a MEMS sensor, which has a low sampling frequency but is much cheaper). In this regard, we propose a transfer-learning model that transfers learning across different instances of manufacturing vibration sensors. This transfer-learning model is based on sharing weights and feature transformations from a deep neural network (DNN) trained with data from the sensor that has a high sampling frequency. These features and weights are used in the classification problem of another sensor data, the one with lower sampling frequency (and thus more sparse data). We show that the transfer-learning idea gives an relative improvement in the accuracy of classifying the defect type of 11.6\% over the regular DNN model. We built variants of DNN models for the defect classification task, i.e., using a single RPM data for training and for testing across the entire operating environment, and using aggregations of data across multiple RPMs for training with interpolation within RPMs.


One may wonder why we need to use sensors with much lower sampling rate; the reason is the significant price difference between the MEMS sensor and piezoelectric sensor. The former has much lower resolution (and also cost \cite{VibrationReport,albarbar2008sensor}---\$8 versus \$1,305). 
Therefore, the goal is to build a predictive maintenance model from the piezoelectric sensor and use it for the MEMS sensor.

In this paper, we test the following hypotheses related to the use of ML models for anomaly detection in IoT-enabled agriculture and manufacturing.

\textbf{Hypothesis 1:} Deep learning-based anomaly detection technique is effective for smart agriculture and smart manufacturing.

\textbf{Hypothesis 2:} The learning process for classifying failures types can be transferred across different sensor types.

Based on our analysis with real data, the results of the above questions are:

\vspace{-3pt}
\begin{enumerate}[noitemsep,topsep=2pt,parsep=0pt,partopsep=2pt,leftmargin=*]

\item We build two time series prediction models for temporal anomaly detection on real sensor data in IoT-based system for detecting anomaly readings collected from the deployed sensors: 1) Long Short-Term Memory (LSTM) algorithm 2) semi-supervised Autoregressive Integrated Moving Average model (ARIMA) model. We test our models for temporal anomaly detection through two real-world datasets which are the data collected from agricultural sensors (e.g., temperature, nitrate concentration, and humidity) and the data collected from manufacturing sensors (e.g., vibration data). The LSTM models outperform the semi-supervised ARIMA models. This is due to the fact that the deployments generate enough data for accurate training of LSTM and due to the complex dependencies among the features of the data set.
 
\item We detect the level of defect (i.e., normal operation, near-failure, failure) for each RPM data using deep learning based classifier (i.e., deep neural network multi-class classifier) and we transfer the learning across different instances of manufacturing sensors. We are the first to bring such idea of transfer learning across different types of sensors. We analyze the different parameters that affect the performance of prediction and classification models, such as the number of epochs, network size, prediction model, failure level, and sensor type. 

\item We make the observation that training at some specific RPMs, for testing under a variety of operating conditions gives better accuracy of defect prediction. The takeaway is that careful selection of training data, namely, by aggregating across multiple of these predictive RPM values, is beneficial. 

\end{enumerate}

The rest of the paper is organized as follows. We first present the anomaly detection and transfer learning models in Section \ref{sec:model}. In Section \ref{sec:anomaly_detect_results}, we present the results of anomaly detection models and analyze the differences in performance between ARIMA and LSTM time-series predictors on agricultural sensors. We report the results on anomaly detection on manufacturing sensors in Section~\ref{sec:anomaly_detect_manuf}. We evaluate the defect-type classification and the transfer-learning model in Section \ref{sec:learning_tranfer_sensors_results}. We present the related work in Section~\ref{sec:lit-review}. Section \ref{sec:conclusion} concludes the paper. 

\section{Proposed Models} \label{sec:model}

\begin{figure*}[h]
\centering
  \includegraphics[width=0.8\textwidth]{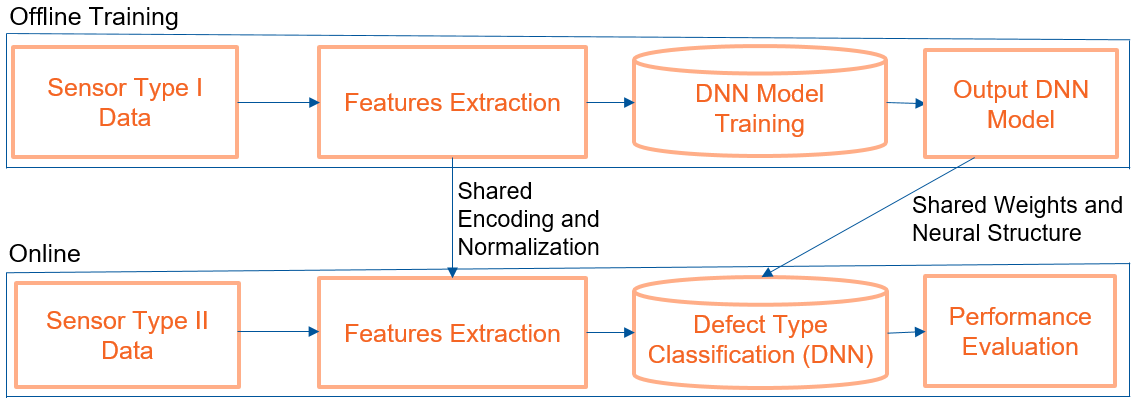}
  \caption{The proposed learning-transfer model is shown. This model has two modes: offline-mode for training the DNN sub-model and online-mode for classifying the sensor under test after sharing knowledge (i.e., DNN's weights and features encoders).}
  \label{fig:learning-transfer model}
  \vspace{6pt}
\end{figure*}

In this section, we describe our proposed algorithms for the anomaly detection and defect type classification.

\subsection{Temporal Anomaly detection:}
Here, we describe our proposed algorithm for detecting anomalies from the sensor readings. First, we build time-series predictors, using two different time-series predictor variants in our  algorithm. We describe them next.

\noindent \textbf{ARIMA models}: 
We build an ARIMA model \cite{contreras2003arima} from the input set of sensor readings. This ARIMA model can be represented as a function \textit{ARIMA(p,d,q)} where $p$ is the number of lag observations included in the regression model, $d$ is the number of times that the raw observations are differenced, and $q$ is the size of the moving average window. Then, we use this trained ARIMA model to detect the anomaly in the sensor's test (future) readings. In practice, $p = 0$ or $q = 0$ as they may cancel each other. We chose $q = 0$, $d = 1$ and $p = 10$ in our model after tuning trials. Specifically, the choice of lag parameter will be explained in detail in the next section. The reason for our choice of ARIMA is that if the data has a moving average linear relation (which we estimated the data does have), ARIMA 
would be better to model such data. Moreover, ARIMA is a simple and computationally efficient model.

\noindent \textbf{LSTM Models}: The second technique is LSTM 
\cite{gers1999learning}, which is a better version than Recurrent Neural Network (RNN) for time-series prediction. It is a well-known RNN architecture that has been used in different applications (e.g., NLP applications~\cite{abdallah2019athena}, classification applications \cite{graves2005framewise}). We train our LSTM variant for time-series prediction since LSTM (with proper choice of the activation function) better models data that has non-linear relationships, which is suitable for manufacturing sensors' readings and, thus, LSTM can be a more expressive model for our anomaly detection task. 

\noindent \textbf{Training Time and Model Performance}: \noindent{Contrasting our two time-series predictor variants:} Although training ARIMA is faster relative to LSTM models (4 hours for semi-supervised ARIMA \textit{vs.} 6 hours for LSTM-based across the 35 sensors in our agriculture deployment), ARIMA still has the requirement of recreating the model for each test sample for improving prediction quality. 
Further, LSTM-based models have better performance relative to ARIMA. Delving into the specifics, the LSTM has lower average RMSE than ARIMA in 6 sensors out of the 7 sensors under test for the five deployed devices in farms.

\noindent \textbf{Anomaly Detection Rule}: 
After using any of the above two proposed time-series predictors, for each sample under test, we would have two values: the actual value (measured by the sensor) and the predicted value (predicted by our model). To flag an anomaly, we consider that $\frac{\text{predicted value} - \text{actual value}}{\text{predicted value}} > 0.2$. In other words, the relative error between the actual value and the predicted value is more than 20\%.  

\subsection{Transfer Learning across Sensor Types:}\label{proposed_model_transfer_learning}
We show our proposed model in Figure~\ref{fig:learning-transfer model} which has two modes: In offline training, the sensor with large amount of data (let us call it sensor type I) has its data entered to the feature extraction module that performs encoding and normalization of the input signals into numerical features. Second, a deep neural network (DNN) model is trained and tuned using these features and labels of the data (where the sample label is normal, near-failure or failure). We use the DNN as a multi-class classifier due to its discriminative power that is leveraged in different classification applications \cite{yuan2016deepgene,ahmed2017unsupervised,al2018computer,elaraby2016deep}. Moreover, DNN is useful for both tasks of learning the level of defect for the same sensor type and for transfer learning across the different sensor types that we consider here.
 
  In online mode, any new sensor data under test (here, sensor type II) would have the same feature extraction process where the saved feature encoders are shared. Then, the classifier predicts the defect type (one of the three states mentioned earlier) given the trained model, and giving as output the probability of each of the three classes. 
 
 It is worth noting that sensor types I and II should be measuring the same physical quantity but can be from different manufacturers and with different characteristics. For instance, in our smart manufacturing domain, sensor type I is a piezoelectric sensor (of high cost  but with high sampling resolution) while type II is a MEMS sensor (of lower cost but with lower sampling resolution). We propose the transfer learning for predictive maintenance i.e., predicting the level of defect of the MEMS sensor and whether it is in normal operation, near-failure (and needs maintenance), or failure (and needs replacement). Another interesting question would be what happens if the two sensor types have overlapping but not identical features in the data that they generate? This may require more complex models which can do feature transformations, using possibly domain knowledge, and we leave such design for future work.

\section{Anomaly Detection with Agricultural Sensors} \label{sec:anomaly_detect_results}

In this section, we use our proposed model to detect the anomaly readings from the sensors. In this context, we evaluate the performance of the model on the real data from our deployed agricultural sensors. 
\subsection{Deployment Details}
 We begin with the agricultural sensors deployment setup, which describes a class of real IoT datasets with a low frequency of data (in our system, every 15 minutes, a new data tuple is saved to our database). For agricultural sensors, we collected the data from 5 different devices at 5 locations (in the Throckmorton Purdue Agricultural Center (TPAC) field) from October 2018 to February 2019. Each location has seven sensors which are: temperature, humidity, soil-conductivity, soil-dielectric, soil-temperature, soil-nitrate, water-nitrate. 
 In this context, the communication between the sensors and the receiver was via LoraWAN \cite{wixted2016evaluation}. We implemented a data collector (server) which is Python-based where receiver collects the data from each sensor on each device every 15 minutes. The sensor data is visualized and available for public experimentation at \cite{Whinsensors}.

\subsection{Experimental Setup} 
 We built two variants of time-series models, semi-supervised ARIMA and LSTM. Each model is a time-series regression model that predicts the value of each sensor. We show the performance of our models in terms of the accuracy of detecting anomalies (measured by mean square error of the model). By evaluating the mean square error, we will study different parameters and setups that affect the performance of the model. The goal of these time-series regression models is to extract the anomaly measures that are typically far from the predicted value of the regression model.
 For each proposed model, the training size was 66\% of the total collected data while the testing size was 34\%. Table~\ref{tbl:size_tranining_arima_agriculture} shows both the training and testing samples of the different sensors. We also varied the proportion of data used for training as a parameter and tested the performance of our model to check the least amount of data needed (which which was 30\% of the data in our experiments) for the time-series regression model to predict acceptable values (within 10\% error from the actual values). 

\begin{table}
\caption {Datasets' description with Sensor Id, Date Deployed , Training Samples, and Testing Samples.} 
\label{tbl:size_tranining_arima_agriculture}
\centering
\small
\begin{tabular}{|c|c|c|c|} 
 \hline
 Sensor Id & Training Samples & Testing Samples \\ [0.5ex] 
 \hline
  22 & 10902 & 5616 \\
  25 & 10634 & 5477 \\
  30 & 7328 & 3775 \\
  32 & 5475 & 2820 \\
  33 & 3955 & 2037 \\
 \hline
\end{tabular}
\end{table}

\subsection{Results and Insights} 

\subsubsection{Semi-supervised ARIMA model:}
\begin{figure}
\centering
  \includegraphics[width=\linewidth]{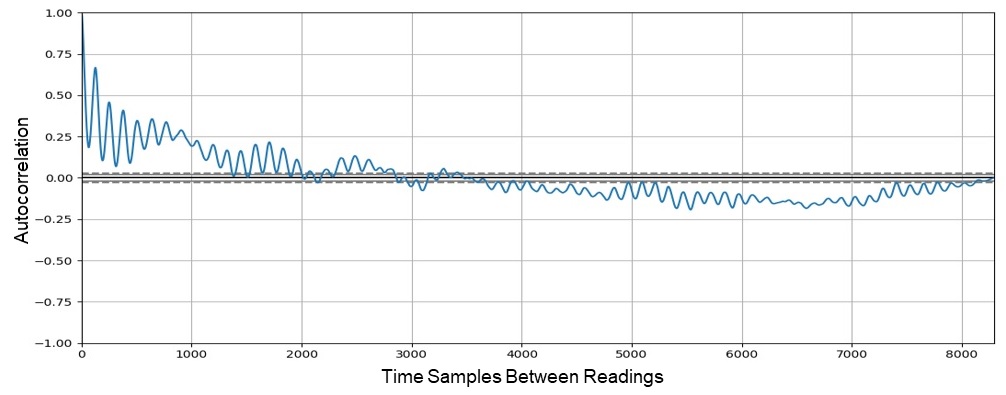}
  \caption{The average correlation between different samples in time from the data collected from agricultural sensors. The correlation is less than 90\% for samples that are separated in time by more than 10 samples. Also, samples that are far away by 2000 have zero or negative correlation. This helps in reducing the size of lag variable in ARIMA regression model which makes the model computationally efficient.}
  \label{fig:correlation_arima_lag}
\end{figure}

As explained earlier, the ARIMA model can be represented as a function \textit{ARIMA(p,d,q)} where $p$ is the number of lag observations included in the regression model, $d$ is the number of times that the raw observations are differenced and $q$ is the size of the moving average window. In this context, the lag-variable $p$ is chosen based on the correlation between the training samples. If two samples are far away from each other, the correlation decreases (i.e., values are almost independent). Thus, the window of lag-variable should be chosen such that for each new fitting of ARIMA model, the older samples that have high dependence should contribute to the next prediction value. Figure~\ref{fig:correlation_arima_lag} shows such fact where 
the correlation is less than 90\% for samples that are separated by more than 10 samples. Also, if we consider samples separated by 2,000 or more time points (where the difference between two consecutive time points is 15 minutes as mentioned earlier), then there is no correlation between those samples. Thus, we chose the lag variable $ p = 10 $ i.e., each new predicted value will be fitted using the 10 last samples. 

For the implementation, we built upon the \texttt{Stats-models} module \cite{seabold2010statsmodels} which is Python-based for creating the two variants of ARIMA models. The first one is the basic ARIMA model and the other one is the semi-supervised ARIMA model that creates a new model for each new sample (the list contains all training data and the new observed sample). The latter (i.e., semi-supervised ARIMA) gave much better performance, thus we use that model for reporting the results and the comparison with LSTM model.

For each type, we build a time-series prediction model and used this model to extract anomaly readings that happen in the collected data. To measure the performance of the semi-supervised ARIMA model, we use the Root Mean Square Error (RMSE). Table~\ref{tbl:arima_agriculture_results} shows both the average test RMSE and number of anomaly measures extracted for each one of the seven types for all sensors. Out of the seven sensors, five have RMSE less than three which shows that the time-series regression model can predict the future value of the sensor effectively. Moreover, the humidity sensor has the least stable time-series regression model due to the weather-humidity change conditions in Indiana in the time period in which the data was collected. Also, we note that the multiple sensors of each kind, specifically Nitrate-based sensors are very close in their RMSEs. Finally, the number of anomalies shows the stability of the data collected from our deployed agricultural sensors.  

\begin{table}
\caption {The average RMSE and number of anomaly samples extracted for each sensor's type for farming data from our deployed sensors using semi-supervised ARIMA model.} 
\centering
\small
\begin{tabular}{|c|c|c|c|} 
 \hline
 Sensor Type &  Average Test RMSE & Number of Anomalies \\ [0.5ex]
 \hline
  Temperature & 1.977 & 3 \\
  Humidity & 5.002 & 4 \\
  Soil-Conductivity & 4.830 & 9\\
  Soil-Dielectric & 2.785 &  2 \\
  Soil-Temperature & 0.627 & 2 \\
  Water-Nitrate & 0.329 & 0 \\
  Soil-Nitrate & 0.076 & 2\\
 \hline
\end{tabular}
\label{tbl:arima_agriculture_results}
\end{table}

\subsubsection{LSTM model results}   
 We used the Long-Short Term Memory (LSTM) which is a better version than Recurrent Neural Network for time-series prediction. We train our LSTM variant to avoid having to make the decision about how to do further tuning, as we have to do for the ARIMA model. We built our model on Keras library \cite{gulli2017deep} which is Python-based. We use 5 epochs for building the model (where more epochs does not give significant enhancements). Same training size and testing size as semi-supervised ARIMA model were used for a fair comparison (note that for semi-supervised ARIMA model, each new sample replaces oldest sample in the history keeping the total size the same).

Table~\ref{tbl:lstm_agriculture_results} shows both the average Test RMSE and number of anomaly measures extracted for each one of the seven types for all sensors. For LSTM models, out of the seven models, six have RMSE less than three.

\begin{table}
\caption {The average Test RMSE, and number of anomaly samples extracted for each sensor's type for farming data from our deployed sensors using LSTM model.} 
\label{tbl:lstm_agriculture_results}
\centering
\small
\begin{tabular}{|c|c|c|} 
 \hline
 Sensor Type &  Average Test RMSE & Number of Anomalies \\ [0.5ex]
 \hline
  Temperature  & 1.39 & 3\\
  Humidity     &  2.71 & 3\\
  Soil-Conductivity & 3.36 & 8\\
  Soil-Dielectric & 1.71 & 3\\
  Soil-Temperature & 0.18 & 2 \\
  Water-Nitrate & 0.53 & 1\\
  Soil-Nitrate & 0.02 & 4\\
 \hline
\end{tabular}
\end{table}

\subsubsection{Comparison between ARIMA and LSTM}
From Tables~\ref{tbl:arima_agriculture_results} and ~\ref{tbl:lstm_agriculture_results}, It is clear that the LSTM model has better performance in predicting the unseen test samples compared to the ARIMA model. Out of the seven types of agricultural sensors, LSTM has better performance in six of the types (i.e., all of the types except water-nitrate sensor type). The reason for such superiority of LSTM is that it better models data that has non-linear relationships (due to the activation function) while ARIMA fits a moving average linear model.
\section{Anomaly detection with Manufacturing Sensors}\label{sec:anomaly_detect_manuf}
The manufacturing sensors are different from agricultural sensors since they have a higher rate of measuring data (e.g., 3.2 KHz for piezoelectric sensor). We build time-series models to predict anomaly in the sensors. 

\subsection{Deployment Details}
\begin{figure*}[t] 
\begin{minipage}[t]{1.0\textwidth}
\begin{minipage}[t]{.32\textwidth}
\centering
\includegraphics[width=0.7\linewidth]{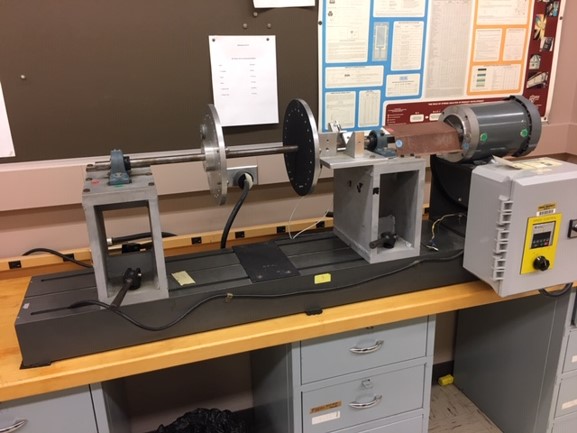}
  \caption{Motor Testbed.}
  \label{fig:testbed1} 
\end{minipage}\hfill
\begin{minipage}[t]{.32\textwidth}
\centering
  \includegraphics[width=0.85\linewidth]{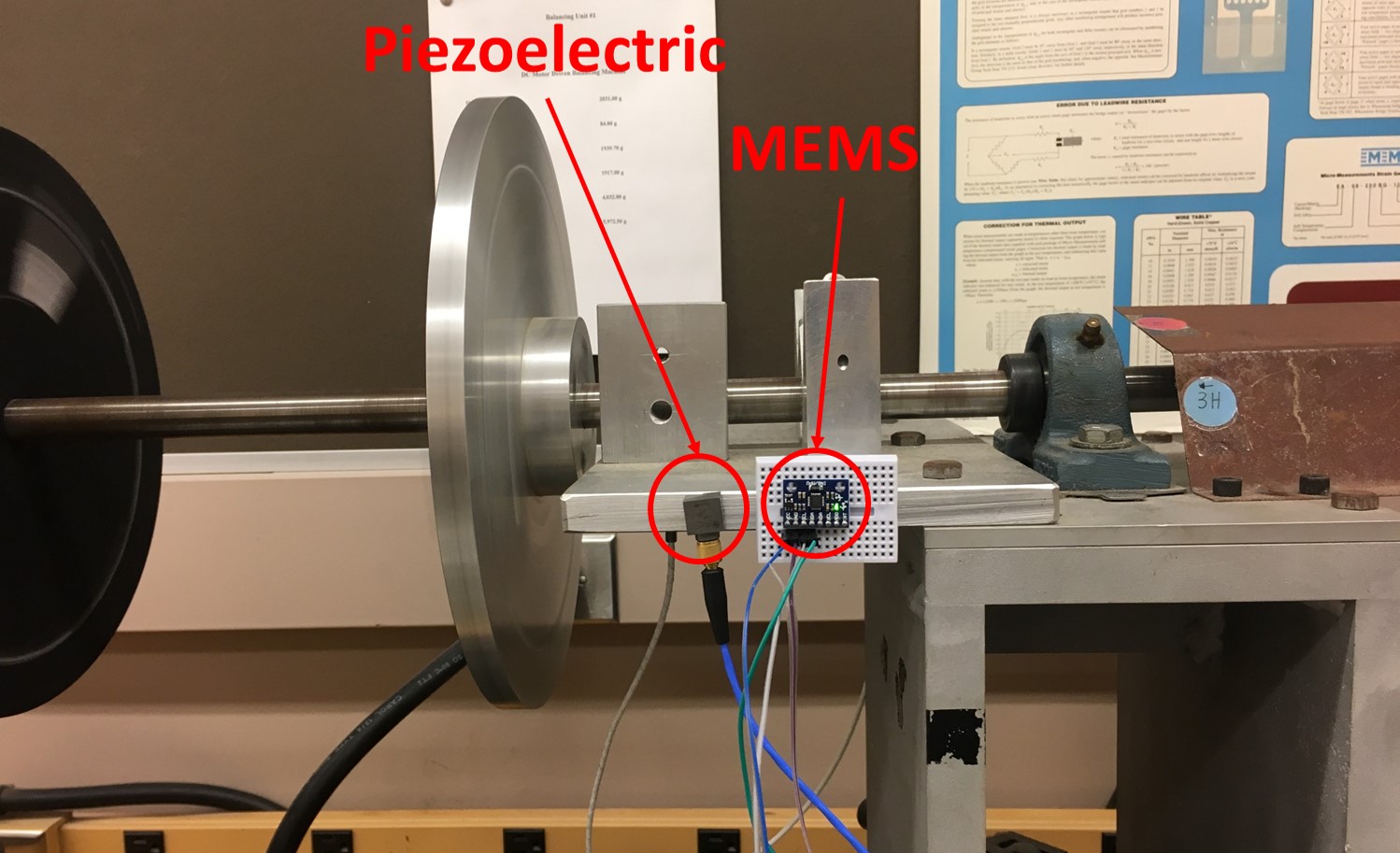}
  \caption{Piezoelectric (left) and MEMS (right) sensors mounted on the motor testbed.}
  \label{fig:testbed2}
\end{minipage}\hfill
\begin{minipage}[t]{.32\textwidth}
\centering
\includegraphics[width=0.7\linewidth]{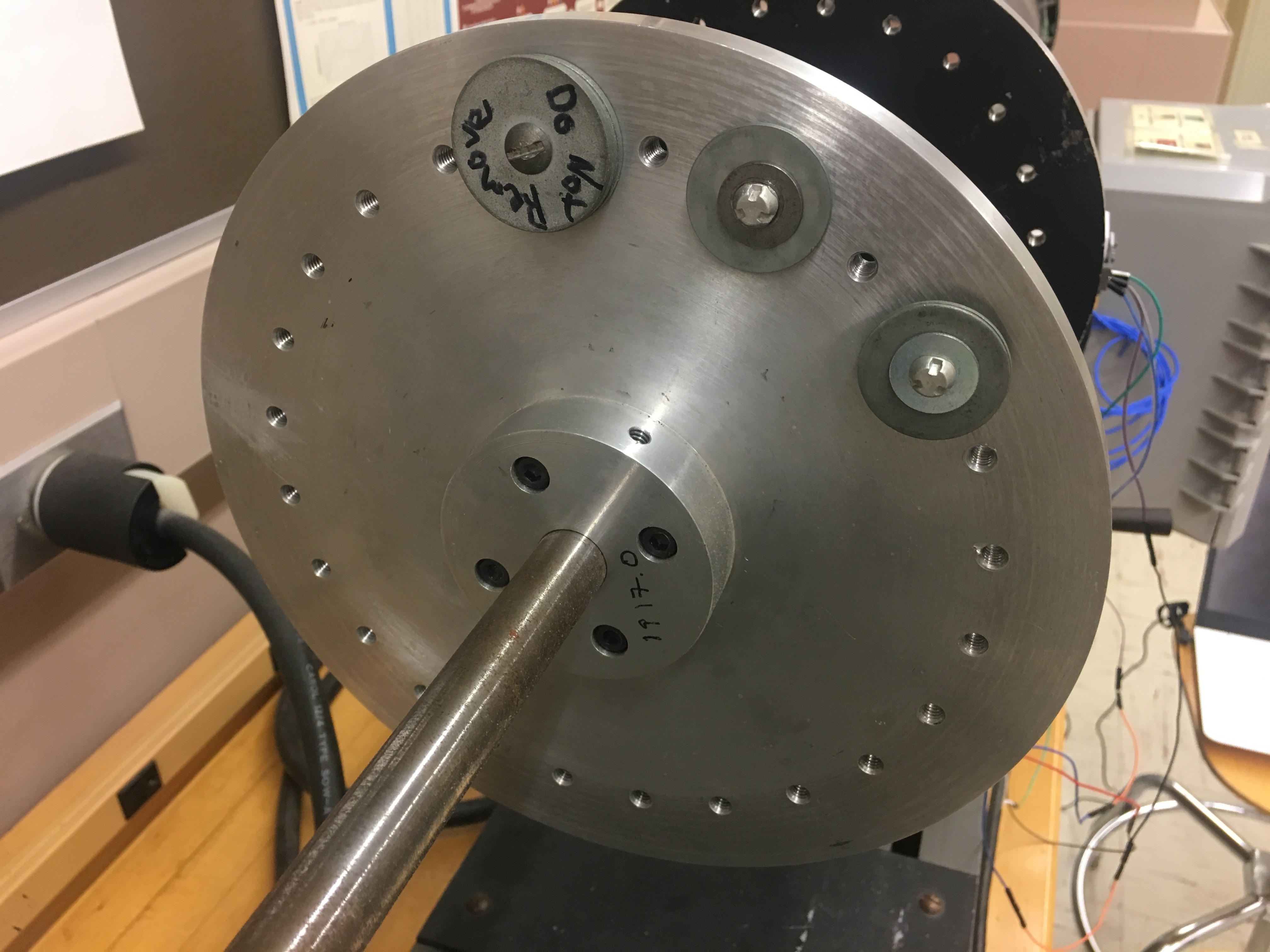}
  \caption{Balancing disk to make different levels of mechanical imbalance in the testbed.}
  \label{fig:testbed3}
\end{minipage}
\end{minipage}
\end{figure*}

An experiment was conducted in the motor testbed (shown in Fig. \ref{fig:testbed1}) to collect machine condition data (i.e., acceleration) for different health conditions. During the experiment, the acceleration signals were collected from piezoelectric and MEMS sensors (shown in Fig. \ref{fig:testbed2}) at the same time with the sampling rate of 3.2 kHz and 10 Hz, respectively, for X, Y, and Z axes. Different levels of machine health condition can be induced by mounting a mass on the balancing disk (shown in Fig. \ref{fig:testbed3}), thus different levels of mechanical imbalance are used to trigger failures. Failure condition can be classified as one of three possible states - normal, near-failure, and failure.
Acceleration data were collected at the ten rotational speeds (100, 200, 300, 320, 340, 360, 380, 400, 500, and 600 RPM) for each condition. While the motor is running, 50 samples were collected at 10 second interval, for each of the ten rotational speeds. We use this same data for defect-type classification and learning transfer tasks in Section~\ref{sec:learning_tranfer_sensors_results}.

\subsection{Experimental Setup} Again, the goal is to measure the performance of our time-series regression model to detect anomalies for the vibration sensors (i.e., on X,Y,Z axes). This performance is measured using RMSE similar to Section~\ref{sec:anomaly_detect_results}.  We trained the ARIMA and LSTM predictive models on specific RPM and tested on another RPM. The data contains different levels of defects (i.e., different labels for indicating normal operation, near-failure, and failure). These labels would be used in next sections. In time-series prediction models, all data that have different levels of defects were tested. Specifically, the data was divided between training and testing equally (i.e., 50\% - 50\%). We stopped after 5 epochs and noticed that after 3 epochs, the total loss on training samples saturates.

\subsection{Results and Insights} 
\subsubsection{LSTM model results}
The time-series prediction works well for all of the vibration sensors. The average RMSE over test data set (50\% of the samples for unseen RPM) was 0.02,0.04,0.02 for X,Y,Z vibration values, respectively. In this context, note that the spectrum of the test time samples captures all three operational states and thus the LSTM model can capture variations efficiently. Also, the same performance was captured for MEMS sensor (figure omitted). 

\subsubsection{Comparison between ARIMA and LSTM} Here, the ARIMA's model performance was worse (RMSE of 0.038, 0.062 ,0.038) due to the non-linear nature of vibration sensors data, the huge number of samples (due to high sampling frequency on piezoelectric sensor).  

\section{Transfer Learning across vibration sensors} \label{sec:learning_tranfer_sensors_results}

In this section, we use our transfer-learning proposed model to detect the level of defect of the readings from the manufacturing sensors. In this context, we evaluate the performance of the model on two real datasets from our manufacturing sensors which are piezoelectric and MEMS vibration sensors. In other words, we perform data analytics on the data from the vibration sensors and infer one of three operational states (mentioned earlier in Section~\ref{sec:anomaly_detect_manuf}) for the motor. 

We show the performance of our model in terms of the accuracy of detecting defect level as measured by the classification accuracy of the deep-learning prediction model on the test dataset which is the defined as the number of correctly classified samples to the total number of samples. We will study different parameters and setups that affect the performance of the model.

The two research questions we answer in this section are:
\begin{itemize}
    \item Can we detect the operational state effectively (i.e., with high accuracy)?
    \item Can we transfer the learned model across the two different types of sensors?
\end{itemize}

\subsection{DNN Model results}

\textbf{Experimental Setup}: We collected the data from 2 deployed sensors, i.e., piezoelectric and MEMS sensors mentioned earlier.  Then, two DNN models were built on these two datasets. First, a normal model for each RPM was built where we train a DNN model on around 480K samples for the RPM. We have a sampling rate of 3.2 KHz (i.e., collect 3.2K data during 1 second) and we collect 50 samples and we have 3 axes. So, total data for one experiment is $3200 x 50 x 3 = 480K$ data points. For testing on same RPM, the training size was 70\% of the total collected data while  the testing size was 30\%. The baseline DNN model consists of 50 neurons per layer, 2 hidden layers (with ReLU activation function for each hidden layer) and output layer with Softmax activation function. Following standard tuning of the model, we created different variants of the models to choose the best parameters (by comparing the performance of the multi-class classification problem). We built upon the Keras library \cite{gulli2017deep} which is Python-based for creating the variants of our models. In our results, we call the two models DNN-R and DNN-TL where the first refer to training DNN regularly and testing on the same sensor while the latter means transfer learning model where training was performed on one sensor and classification was performed on a different sensor (using the design of shared weights and learned representations as described in Sec.~\ref{proposed_model_transfer_learning}). Specifically, for the DNN-TL, training was done on the plentiful sensor data from the piezoelectric sensor and the prediction was done based on the MEMS sensor data. The comparison between regular DNN model and our transfer-learning DNN model on MEMS sensors in terms of the best achieved accuracy is shown in Table~\ref{tbl:comparison_results_dnn_tl}. We notice that the transfer-learning model gives a relative gain of 11.6\% over the model trained only on the lower resolution MEMS sensor data. The intuition here is that the MEMS sensor data is only 2000 samples, due to very low sampling rate (10 Hz as opposed to 3.2 kHz with the piezoelectric sensor) and thus it cannot fit a good DNN-R model.  On the other hand, we can train a DNN-TL model with sensor of different type (but still with vibration readings) with huge data and classify the failure of the sensor under test (i.e., MEMS with less data) with accuracy 71.71\%. 

\begin{table}
\caption {A comparison between regular DNN model and our transfer-learning DNN model on MEMS sensors. The transfer-learning model gives an absolute gain of 7.48\% over regulard DNN model.} 
\label{tbl:comparison_results_dnn_tl}
\centering
\small
\begin{tabular}{|c|c|c|} 
 \hline
 Model Type &  Sensor Tested & Accuracy (\%) \\ [0.5ex] 
 \hline
  DNN-R & MEMS & 64.23\% \\
  DNN-TL & MEMS &  71.71\% \\
  DNN-R & piezoelectric &  80.01\% \\
 \hline
\end{tabular}
\end{table}

Moreover, we show the effect of parameter-tuning on the performance of the models in Table~\ref{tbl:dnn_tuning_results}. The parameter tuning gives relative gain of 11.91\% over the baseline DNN-TL model. Delving into the specifics, the most effective tuning steps were feature-normalization which gives relative increase of 7.15\% in the accuracy over non-normalized features and increasing number of hidden layers and batch size which gave around 1.36\% each on the performance. Note that increasing the epochs to 200 and hidden layers to more than 3 decreases the accuracy, due to over-fitting. 

\begin{table}%
\caption {The effect of parameter tuning on the accuracy of DNN-TL model. Here training is done on the plentiful sensor data from the piezoelectric sensor and testing is done on the low sampling rate MEMS sensor data. The parameter tuning gave an absolute gain of 7.63\% over the baseline model.} 
\label{tbl:dnn_tuning_results}
\centering
\small
\begin{tabular}{|c|c|} 
 \hline
  Tuning Factor & Accuracy \\ [0.5ex]
 \hline
   None & 58.00\% \\
  Feature Selection & 64.08\% \\
  Feature-normalization & 68.66\% \\
  Neurons per layer (50-80-100) & 69.41\% \\
  Number of Hidden layers (2-3) & 70.32\% \\
  Number of Epochs (50-100) & 70.75\% \\
  Batch Size (50-100) & \textbf{71.71\%} \\
 \hline
\end{tabular}
\vspace{-5pt}
\end{table}

\textbf{Feature Selection:} We validate one idea that the vibration data in certain axis will not carry different information in normal and failure cases. The circular movement around the center of the motor is on X and Z axes so that they have vibration values that change with motor condition while the Y-axis has smaller vibration (the direction of the shaft). Thus, we compare the result of the model when the features are the three data axes in one setup (i.e., default setup) and the proposed idea when the features are extracted only from X-axis and Z-axis data vectors.
According to the experimental setup shown in Fig.~\ref{fig:testbed1}, as the motor rotates with the disk, which is imbalanced by the mounted mass, i.e., eccentric weight, the centripetal forces become unbalanced, and this causes repeated vibrations along multiple directions. Considering the circular movement around the center of the motor, the two directions, which are x-axis and z-axis in our case, are mainly vibrated while the y-axis (the direction along the shaft) show relatively smaller vibration, which may not show a distinguishable variation in the data pattern as machine health varies. We find that this feature selection process gave us a relative increase of 10.5\%  over the baseline model with all three features. Specifically, the accuracy is 58\% using the  model trained on default features compared to 64.08\% using the model with feature selection. This kind of feature selection requires domain knowledge, specifically about the way the motor vibrates and the relative placement of the sensors.
The intuition here is that redundant data features are affecting the model's learning and therefore selecting the most discriminating features helps the neural network learning. 


\subsection{Data-augmentation model results}   
\textbf{Experimental Setup}: We used data-augmentation techniques (by both augmenting data from different RPMs and generating samples with interpolation within each RPM) and train DNN-R model on each sensor. For piezoelectric sensor, the data-augmentation model consists of 5M samples (480K samples collected from each rotational speed data for the available ten rotational speeds and 20K generated samples by interpolation within each RPM). For MEMS sensor, the data augmentation model consists of 15120 samples. We compare the average accuracy of the model over all RPMs under the regular model (DNN-R) and the augmented model. The absolute increase in the accuracy using the augmentation techniques over the regular model is 9.76\% for piezoelectric and 8.99\% for MEMS, respectively. The intuition here is that our 
data-augmentation techniques are useful for both piezoelectric and MEMS vibration sensors. Data-augmentation is useful for transfer learning across different RPMs. 

\textbf{Effect of variation of RPMs results}: 
Here, we show the details of each RPM-single model and the details of the data-augmented model. First, we train a single-RPM model and test that model on all RPMs. Then, we build a data-augmented model as explained earlier. Table~\ref{tbl:different_rpms_results} shows such comparison where the single-RPM model can't transfer the knowledge to another RPMs. An interesting note is that at the slowest RPMs (here, RPM-100 and RPM-200) the separation is harder at the boundary between failure, near-failure, and normal operational states. On the other hand, data-augmented model has such merit since it is trained on different samples from all RPMs with adding data-augmentation techniques. In details, the absolute enhancement in the average accuracy across all RPMs is 6\% while it is 13\% over the worst single-RPM model (i.e., RPM-600). 
In the data-augmented model, 70\% from each RPM's samples were selected for training that model as mentioned earlier.

\begin{table}
\Large
\caption {Comparison on the performance of failure detection model where the trained model is using one RPM and the tested data is from another RPM. The data-augmentation model is useful for transfer the learning across different RPMs. The absolute enhancement in the average accuracy across all RPMs is 6\% while it is 13\% over the worst single-RPM model.} 
\label{tbl:different_rpms_results}
\centering
\resizebox{0.48\textwidth}{!}{
\begin{tabular}{|c|c|c|c|c|c|c|c|} 
 \hline
 Trained RPM &  RPM-100 & RPM-200 & RPM-300 & RPM-400 & RPM-500 & RPM-600 & Average (\%) \\ [0.5ex]
 \hline
  RPM-100 & 68.80\% & 66.64\% & 65.83\% & 73.61\% & 67.29\% &  42.90\% & 64.18\% \\
  RPM-200 & 63.54\% & 73.71\% & 58.11\% & 74.67\% & 67.68\% &  45.93\% & 63.94\% \\
  RPM-300 & 57.99\% & 55.00\% & 95.20\% &  66.09\% & 71.32\%    &   45.14\%	&     65.12\%  \\
  RPM-400 & 66.37\% & 69.68\% & 54.79\% &  87.38\% & 69.52\%    &  32.62\% & 63.39\%     \\
  RPM-500 & 65.37\% &  64.94\% & 80.12\% & 80.20\% & 75.61\% & 42.59\% & 68.06\%  \\
  RPM-600 & 49.16\% & 51.12\% &  63.44\% & 44.23\% &  55.02\% & 75.16\% & 56.35\%   \\
  Augmented-data model & 67.94\% & 71.31\%  & 62.61\%  & 80.06\% & 69.06\% & 65.88\% & \textbf{69.48\%}  \\
 \hline
\end{tabular}
}
\vspace{-5pt}
\end{table}

\textbf{Confusion Matrices Comparison}: Here, we show the confusion matrix which compare the performance of our DNN-R models for each operational stateseparately. Table~\ref{tbl:confusion_matrix_aug} shows such metric using data-augmentation. The best performance is for near-failure which exceeds 96\%. This is good in practice since it gives early alarm (or warning) about the expected failure in future. Moreover, the model has good performance in normal operation which exceeds 70\%. Finally, the failure accuracy is a little lower which is 61.67\% however the confusion is with near-failure state which also gives alarm under such prediction. On the other hand, DNN-R model without data-augmentation has worse prediction in both normal and near-failure modes as shown in Table\ref{tbl:confusion_matrix_noaug} (normal operation detection around 60.4\% and near-failure is 93.86\%) while much better for detecting failures where the accuracy is 75.00\%. The intuition here is that detecting near-failure and normal-operation modes can be enhanced using data-augmentation techniques. On the contra, detecting failure operational state is better without data-augmentation as failure nature can be specific for each RPM and thus creating single model for each RPM can be useful in that sense.

\begin{table}
\caption{Confusion matrices for the classification of operational conditions using DNN-R and data-augmentation.}
\label{tbl:confusion_matrix_aug}
\begin{tabular}{l|l|c|c|c|}%
\multicolumn{2}{c}{}&\multicolumn{2}{c}{Predicted Mode}\\
\cline{3-5}
\multicolumn{2}{c|}{}& Normal & Near-failure & Failure\\
\cline{2-5}
& Normal & $\textbf{70.22\%}$ & $28.40\%$ & $1.38\%$\\ 
\cline{2-5}
& Near-failure & $3.82\%$ & $\textbf{96.05\%}$ & $0.13\%$\\
\cline{2-5}
& Failure & $0\%$ & $38.33\%$ & $\textbf{61.67\%}$\\
\cline{2-5}
\end{tabular}%
\end{table}

\begin{table}[h]
\caption{Confusion matrix for the classification of operational conditions using DNN-R and no data augmentation.}
\label{tbl:confusion_matrix_noaug}
\begin{tabular}{l|l|c|c|c|}
\multicolumn{2}{c}{}&\multicolumn{2}{c}{Predicted Mode}\\
\cline{3-5}
\multicolumn{2}{c|}{}& Normal & Near-failure & Failure\\
\cline{2-5}
& Normal & $\textbf{60.38\%}$ & $18.52\%$ & $21.09\%$\\
\cline{2-5}
& Near-failure & $6.08\%$ & $\textbf{93.86\%}$ & $0.06\%$\\
\cline{2-5}
& Failure & $0\%$ & $25.00\%$ & $\textbf{75.00\%}$\\
\cline{2-5}
\end{tabular}
\end{table}

\subsection{Relaxation of classification problem:}
In some applications of the sensor data, the goal can be to detect only if the data from the deployed sensor is normal or not. Thus, we relax the defect classification problem into binary classification problem to test such application. 
In this subsection, the experimental data obtained under five rotation speeds, i.e., 300, 320, 340, 360, and 380 RPMs were considered to classify between normal and not-normal states. For the deep learning model, we use neural network, which consists of two layers. 

The models’ performances are summarized in Table \ref{tbl:binary_CNN_different_rpms_results}. Compared to Table~\ref{tbl:different_rpms_results}, the performance here is better due to the following reasons. First, the confusion is less in binary classification problem since we have only two classes. Second, the variation in the range between RPMS is less in this experiment. 
 \begin{table}[htbp]
 \caption {Comparison on the performance of binary classifier detection model where the trained model is using one RPM and the tested data is from another RPM.  The average accuracy is higher, compared to the three classes defect classification models.} 
\label{tbl:binary_CNN_different_rpms_results}
\centering
\resizebox{0.48\textwidth}{!}{
\begin{tabular}{|c|c|c|c|c|c|c|} 
 \hline
 Trained RPM &  RPM-300 & RPM-320 & RPM-340 & RPM-360 & RPM-380 &  Average (\%) \\ [0.5ex]
 \hline
  RPM-300 & \textbf{100\%} & 65.17\% & 58.17\% & 51.50\% &  50.63\% & 65.09\% \\
  RPM-320 & 99.75\% & \textbf{100\%} & 97.58\% & 78.63\% & 68.17\% &  88.82\% \\
  RPM-340 & 96.60\% & 99.27\% & \textbf{100\%} &  97.33\% & 82.43\%    &  95.12\%  \\
  RPM-360 & 96.60\% & 99.27\% & 97.33\% &  \textbf{99.67\%} & 84.43\% & \textbf{95.46\%} \\
  RPM-380 & 61.05\% &  87.75\% & 96.93\% & 99.83\% &  \textbf{99.77\%} & 89.07\%  \\
 \hline
\end{tabular}
}
\end{table}

\section{Related Work}\label{sec:lit-review}

\noindent {\bf Failure detection Models}:
There have been several works to study a failure detection in manufacturing processes using single or multi-sensor data \cite{lee2019mfg2,teng1996failure,lee2019mfg3}. Specifically, the recent work \cite{lee2019mfg2}, in which the kernel principal component analysis (KPCA) based anomaly detection system was proposed to detect a cutting tool failure in a machining process. In the study, multi-sensor signals were used to estimate the condition of a cutting tool, but a transfer learning between different sensor types was not considered. Also, in another recent study \cite{lee2019mfg3}, the fault detection monitoring system was proposed to detect   various failures in a DC motor  such as a gear defect, misalignment, and looseness. In the study, a single sensor, i.e., accelerometer, was used to obtain machine condition data, and several convolutional neural network architectures were used to detect the targeted failures. However, different rotational speeds and sensors were not considered. Thus, these techniques must be applied again for each new sensor type. On the other hand, we consider the transfer learning between different sensor types.

\noindent {\bf Learning Transfer}: 
Transfer learning is the improvement of learning in a new task through the transfer of knowledge from a related task that has already been learned \cite{torrey2010transfer}. Transfer learning has been proposed to allow the domains, tasks, and distributions to be different, which can extract knowledge from one or more source tasks and apply the knowledge to a target task \cite{xiang2010bridging,qiu2016survey} with the advantage of intelligently applying knowledge learned previously to
solve new problems faster. In the literature, transfer learning techniques have been applied successfully in many real-world data processing applications, such as cross-domain text classification, constructing informative priors, and large-scale document classification \cite{ling2008spectral,chen2015net2net}. However, these works did not tackle the transfer the learning across different instances of sensors that we consider here.

\section{Conclusion}\label{sec:conclusion}

We studied \emph{anomaly detection} and \emph{failure classification} for the predictive maintenance problem of IoT-based digital agriculture and smart manufacturing. We designed a temporal anomaly detection technique and an efficient defect-type classification technique for these two application domains. We compared the strategies of LSTM and semi-supervised ARIMA for anomaly detection. We observed that LSTM leads to better anomaly detection prediction at the cost of longer training time. We tested our findings on two real-world data-sets. We also studied the effects of several tuning parameters to enhance the failure classification 

We proposed a transfer learning model for classifying failure on sensors with lower sampling rate (MEMs) using learning from sensors with huge data (piezoelectric). Our findings indicate that the transfer learning model can considerably increase the accuracy of failure detection. Finally, we use data-augmentation techniques to enhance the prediction of the failure mode. Using such augmentation, the accuracy was enhanced with the enhancement becoming more pronounced in near-failure mode. 

As a future work, correlating and leveraging the data from multiple sensors is one possible extension. Moreover, detecting the device health by merging information from multiple, potentially different, sensors will be explored. 

\begin{acks}
This work is supported by the Wabash Heartland Innovation Netowrk (WHIN). The opinions expressed in this publication are those of the authors. They do not purport to reflect the opinions or views of sponsor.
\end{acks}

\bibliographystyle{ACM-Reference-Format}
\bibliography{sample}



\end{document}